\documentclass[conference]{IEEEtran}

\usepackage{cite}

\usepackage[pdftex]{graphicx}

\usepackage[cmex10]{amsmath}
\usepackage{amsfonts}
\usepackage{flushend}
\usepackage{amsthm}

\usepackage{algorithm}
\usepackage{algorithmicx}
\usepackage{algpseudocode}
\usepackage{xspace}

\usepackage{multirow}
\usepackage[table]{xcolor}
\usepackage{colortbl}
\definecolor{res2}{RGB}{220,220,220}
\definecolor{res}{RGB}{180,180,180}

\newcommand{\omf}{\textsc{OneMax}\xspace}
\newcommand{\zmf}{\textsc{ZeroMax}\xspace}
\newcommand{\omdf}{\textsc{$OM_d$}\xspace}
\graphicspath{{pic/}}

\allowdisplaybreaks

\IEEEoverridecommandlockouts

\begin{document}

\title{Reinforcement Learning Based \\ Dynamic Selection of Auxiliary Objectives \\ with Preserving of the Best Found Solution}
\author{
\IEEEauthorblockN{Irina Petrova, Arina Buzdalova}
\IEEEauthorblockA{
ITMO University\\
49 Kronverkskiy av.\\
Saint-Petersburg, Russia, 197101\\
Email: irenepetrova@yandex.com, abuzdalova@gmail.com}
}

\maketitle

\begin{abstract}
Efficiency of single-objective optimization can be improved by introducing some auxiliary objectives. Ideally, auxiliary objectives should be helpful. However, 
in practice, objectives may be efficient on some optimization stages but obstructive on others. In this paper we propose a modification of the EA+RL method 
which dynamically selects optimized objectives using reinforcement learning. The proposed modification prevents from losing the best found solution.

We analysed the proposed modification and compared it with the EA+RL method and Random Local Search on XdivK, Generalized OneMax and LeadingOnes problems. 
The proposed modification outperforms the EA+RL method on all problem instances. It also outperforms the single objective approach on the most problem 
instances. We also provide detailed analysis of how different components of the considered algorithms influence efficiency of optimization. In addition, we 
present 
theoretical analysis of the proposed modification on the XdivK problem.

\end{abstract}

\section{Introduction}

Consider single-objective optimization of a \emph{target} objective by an evolutionary algorithm (EA). Commonly, efficiency of EA is measured in one of two 
ways. In the first one efficiency is defined as the number of fitness function evaluations needed to reach the optimum. In the second one efficiency of EA is 
computed as the maximum target objective value obtained within the fixed number of evaluations. In this work we use the first way.

Efficiency of the target objective optimization 
can be increased by introducing some \emph{auxiliary} objectives~\cite{coello, mh-iff, neumann, brockhoff, mz-theory3}. Ideally, auxiliary objectives 
should be helpful~\cite{mh-iff}. However, in practice, 
objectives can be generated automatically and may be efficient on some optimization 
stages but obstructive on others~\cite{earl-cec, job-shop-later}. We call such objectives \emph{non-stationary}. One of the approaches to deal with such 
objectives is dynamic 
selection of the best objective at the current stage of optimization. 
The objectives may be selected randomly~\cite{helpers}. The better method is EA+RL which uses reinforcement learning~(RL)~\cite{sutton, aux}.

It was theoretically shown for a number of optimization problems that EA+RL efficiently works with stationary objectives~\cite{xdivk, 
leadingones-onemax-cec15}. However, theoretical analysis of EA+RL with non-stationary objectives 
showed that EA+RL does not ignore obstructive objective on the XdivK problem~\cite{petrova-mendel2016}. Selection of an inefficient auxiliary objective causes 
losing 
of 
the best 
found solution and the algorithm needs a lot of steps to find a good solution again. Also EA+RL can stuck in local optima while solving the Generalized OneMax 
problem with obstructive objectives~\cite{buzdalovaPB-conflicting}. 

In this paper we propose a modified version of EA+RL and analyse it theoretically and experimentally on XdivK, Generalized OneMax and LeadingOnes problems.
The rest of the paper is organized as follows. First, the EA+RL and model problems with non-stationary objectives are described. Second, a modification of the 
EA+RL method is proposed. Then we experimentally analyse the considered methods. Finally, we provide discussion and theoretical explanation of the achieved 
results.

\section{Preliminaries}

In this section we describe the EA+RL method. Also we define model problems and non-stationary objectives used in this study. 

\subsection{EA+RL method}
In reinforcement learning (RL) an \emph{agent} applies an \emph{action} to an \emph{environment}. Then the environment returns a numerical reward and a
representation 
of its \emph{state} and the process repeats. The goal of the agent is to maximize the total reward~\cite{sutton}. 

In the EA+RL method, EA is treated as an environment, selection of an objective to be optimized corresponds to an action. 
The agent selects an objective, EA generates new population using this objective and returns some reward to the agent. The reward depends on 
difference of the 
best target objective value in two subsequent generations. We consider maximization problems. So the higher is the newly obtained target value, the higher is 
the reward.

In recent theoretical analysis of EA+RL with non-stationary objectives, random local 
search 
(RLS) is used instead of EA, population consists of a single individual~\cite{petrova-mendel2016}. Individuals are represented as bit strings, flip-one-bit 
mutation is used. If values of the selected objective computed on the new individual and on the current one are equal, the new individual is 
accepted. The used reinforcement learning algorithm is Q-learning. Therefore, the EA+RL 
algorithm in this case is 
called \emph{RLS~+~Q-learning}. The pseudocode of RLS~+~Q-learning is presented in Algorithm~\ref{pseudo}. The current state $s$ is defined as the target 
objective value of the current individual. The reward is calculated as difference of the target objective values in two subsequent generations. In 
Q-learning, the efficiency of 
selecting an objective $h$ in a state $s$ is measured by the value $Q(s,a),$ which is updated dynamically after each selection as shown in line~\ref{q-formula} 
of the pseudocode, where $\alpha$ and $\gamma$ are the learning rate and the discount factor correspondingly.

	\begin{algorithm}[h!]
		\caption{RLS~+~Q-learning Algorithm}
		\label{pseudo}
		\begin{algorithmic}[1]
			\State {Individual $y \gets$ a random bit string}
			\State {Construct set $H$ of auxiliary objectives and target objective} 
			\State {$Q(s,h)$ $\gets 0$ for each state $s$ and objective $h \in H$}
			\While {(Optimum of the target objective $t$ is not found)}
				\State {Current state $s$ $\gets$ $t(y)$}
				\State {Individual $y'$ $\leftarrow$ mutate $y$ (flip a random bit)}
				\State {Objective $h$: $Q(s, h)=\max_{h' \in H} Q(s, h')$}
				\Comment{If Q-values are equal, objectives are selected equiprobably}
				\If{$h(y') \geq h(y)$}\State{$y \gets y'$}\EndIf
				\State {New state $s'$ $\gets t(y)$}
				\State{Reward $r$ $\gets$ $s' - s$}
				\State{$Q(s,h)$ $\gets$ $Q(s,h)+\alpha(r + \max \limits_{h' \in H}{Q(s',h')} - Q(s, h))$} \label{q-formula}
			\EndWhile
		\end{algorithmic}
	\end{algorithm}

\subsection{Model problems} \label{modelproblems}
In this paper we consider three model problems which were used in studies of EA+RL~\cite{petrova-mendel2016, 
leadingones-onemax-cec15, buzdalovaPB-conflicting}.  
In all considered problems, an individual is a bit string of length $n$. 
Let $x$ be the number of bits in an individual which are set to one. 
Then the objective \omf is equal to $x$ and the objective \zmf is equal to $n - x$. 

One of the considered problems is \textsc{Generalized OneMax}, denoted as \omdf. The target objective of this problem called \omdf is calculated as the number 
of 
bits in an 
individual of 
length $n$ that matches a given bit mask. The bit mask has $d$ 0-bits and $n-d$ 1-bits.

Another problem is \textsc{XdivK}. The target objective is calculated as $\lfloor \frac{x}{k} \rfloor$, where $x$ is the number of ones, $k$ is a constant, 
$k < n$ and $k$ divides $n$. 

The last considered problem is \textsc{LeadingOnes}. The target objective of this problem is equal to the length of the maximal prefix consisting of bits set 
to one. 

\subsection{Non-stationary objectives}
We used the two following non-stationary auxiliary objectives for all the considered problems. These auxiliary objectives can be both \textsc{OneMax} or 
\textsc{ZeroMax} at different stages of optimization.
More precisely, consider the auxiliary objectives $h_1$ and $h_2$ defined in~\eqref{h}. The parameter $p$ is called a \emph{switch point}, because at 
this point auxiliary objectives change their properties. \omf and \zmf are denoted as \textsc{OM} and \textsc{ZM} correspondingly.
\begin{equation} 
\hspace{-0.5em}
 h_1(x) =
  \begin{cases}
   \textsc{OM}, x \leq p\\
   \textsc{ZM}, p < x \leq n
  \end{cases} 
  h_2(x) =
  \begin{cases}
   \textsc{ZM}, x \leq p\\
   \textsc{OM}, p < x \leq n
  \end{cases} \hspace{-1.5em} \label{h}
\end{equation}

In the \textsc{XdivK} problem, an objective which is currently equal to \textsc{OneMax} allows to distinguish individuals with the same value of the target 
objective and give preference to the individual with a higher $x$ value. Such an individual is more likely to produce a descendant with a higher target 
objective 
value.
In the \textsc{LeadingOnes} problem, \omf is helpful because it has the same optimum but running time of optimizing \omf is 
lower~\cite{augerdoerr}.
Therefore, in \textsc{LeadingOnes} and \textsc{XdivK} problems the objective which is equal to \omf at the current stage of optimization is 
helpful. In both these problems, the goal is to obtain individual of all ones, so \zmf is obstructive.

In the \omdf problem, both objectives may be obstructive or neutral. For example, if for some $i$ the $i$-th bit of the \omdf bit mask is set to 1, and the 
mutation 
operator flips the $i$-th bit of an individual from 0 to 1, the \zmf objective is obstructive, because it would not accept this individual. 
In the inverse case, \omf is obstructive. 

\section{Modified EA+RL}\label{defs}
In this section we propose a modification of the EA+RL method which prevents EA+RL from losing the best found solution.
In the EA+RL method, if the newly generated individual is better than the existing one according to the selected objective, the 
new individual is accepted. However, if the selected objective is obstructive, the new individual may be worse than the existing individual in terms of the 
target 
objective. In this case EA loses the individual with the best target objective value. 

In the modified EA+RL, if the newly generated individual is better than the existing one according to the selected objective, but is worse according to the 
target objective, the 
new individual is rejected. As in the recent theoretical works, we use RLS as optimization problem solver and apply Q-learning to select objectives. The 
pseudocode of the modified RLS~+~Q-learning is presented in Algorithm~\ref{newpseudo}. 

\begin{algorithm}[h!]
		\caption{Modified RLS~+~Q-learning Algorithm}
		\label{newpseudo} 
		\begin{algorithmic}[1]
			\State {Individual $y \gets$ a random bit string}
			\State {Construct set $H$ of auxiliary objectives and  target objective} 
			\State {$Q(s,h)$ $\gets 0$ for each state $s$ and objective $h \in H$}
			\While {(Optimum of the target objective $t$ is not found)}
				\State {Calculate current state $s$}
				\State {Save target fitness: $f \gets t(y)$}
				\State {Individual $y'$ $\leftarrow$ mutate $y$ (flip a random bit)}
				\State {Objective $h$: $Q(s, h)=\max_{h' \in H} Q(s, h')$}
				\Comment{If Q-values are equal, objectives are selected equiprobably}
				\If{$h(y') \geq h(y)$ and $t(y') \geq t(y)$} 
				  \State{$y \gets y'$}  
				\EndIf
				\State {Calculate new state $s'$} \label{statecalc}
				\State{Calculate reward $r$} \label{rewardcalc}
				\State{$Q(s,h)$ $\gets$ $Q(s,h)+\alpha(r + \max \limits_{h' \in H}{Q(s',h')} - Q(s, h))$}
			\EndWhile
		\end{algorithmic}
	\end{algorithm}
	
To motivate the approach of reward calculation in the new method, we need to describe how the agent learns which objective 
should be selected in the EA+RL method. If the agent selects an obstructive objective and EA loses individual with the best target value, the best target value 
in the new generation is decreased. So the agent achieves a negative reward for this objective and will not select this objective in the same state 
later. However, when properties of auxiliary objectives are changed, the obstructive objective may become helpful. If the properties are changed within 
one 
RL state, the objective which became helpful will not be selected, because the agent previously achieved negative reward for it. And inversely, the 
objective which became obstructive could be selected because it was helpful earlier. And, as it was shown in~\cite{petrova-mendel2016}, 
the 
EA+RL needs a lot of steps to get out of this trap. For this reason, in the new method we consider two versions of the modified EA+RL with different ways of 
reward calculation.

In the first version of the modified EA+RL, if the new individual 
is better than the current one according to the selected objective, and its target objective value is lower, the new individual is rejected but the agent 
achieves negative 
reward as if the new individual was accepted. So the agent learns as in EA+RL. We call this algorithm \emph{the modification of EA+RL with learning on 
mistakes}. In this case in 
line~\ref{rewardcalc} of Algorithm~\ref{newpseudo} reward is calculated as presented in Algorithm~\ref{reward}. In the second version the agent in the same 
situation achieves zero reward 
because the individual in the population is not changed. Thereby, agent does not learn if the action was inefficient and learns only if the target objective 
value was 
increased. We call this algorithm \emph{the modification of EA+RL without learning on mistakes}. In this case in line~\ref{rewardcalc} of 
Algorithm~\ref{newpseudo} reward is calculated as $t(y) - f$.

\begin{algorithm}[h!]
		\caption{Reward calculation in the modified EA+RL with learning on mistakes (Algorithm~\ref{newpseudo}, line~\ref{rewardcalc})}
		\label{reward}
		\begin{algorithmic}[1]
		        \State {$r \gets t(y) - f$}
				\If{$h(y') \geq h(y)$}
				  \State{$r \gets t(y') - f$}   
				\EndIf
		\end{algorithmic}
	\end{algorithm}

In the existing theoretical works on EA+RL, RL state is defined as the target objective value~\cite{xdivk, buzdalovaPB-conflicting}. Denote it as \emph{target} 
state. However, if the individual with the best target objective value is preserved, the algorithm will never return to the state where it achieved positive 
reward. So the agent never knows which objective is helpful. It only can learn that an objective is obstructive if the agent achieved negative reward for 
it. Therefore, in the present work we consider two definitions of a state. The first one is the target state. The second one is the \emph{single} state which 
is the same in the 
whole optimization process. This state is used to investigate efficiency of the proposed method when the agent has 
learned which objective is good.

\section{Theoretical analysis}
Previously, it was shown that the EA+RL method gets stuck in local optima on XdivK with non-stationary objectives~\cite{petrova-mendel2016}. Below we present 
theoretical analysis of the running time of the proposed EA+RL modification without learning on mistakes on this problem. The target state is used.
To compute the expected running time of the algorithm, we construct the Markov chain that represents the corresponding optimization 
process~\cite{xdivk, petrova-mendel2016}. 
Recall that RL states are determined by the target objective value. Markov states correspond to the number of 1-bits in an individual. Therefore, 
an RL state includes $k$ Markov states with different number of 1-bits.

\begin{figure}[h]
\begin{center}
\includegraphics[scale=0.8]{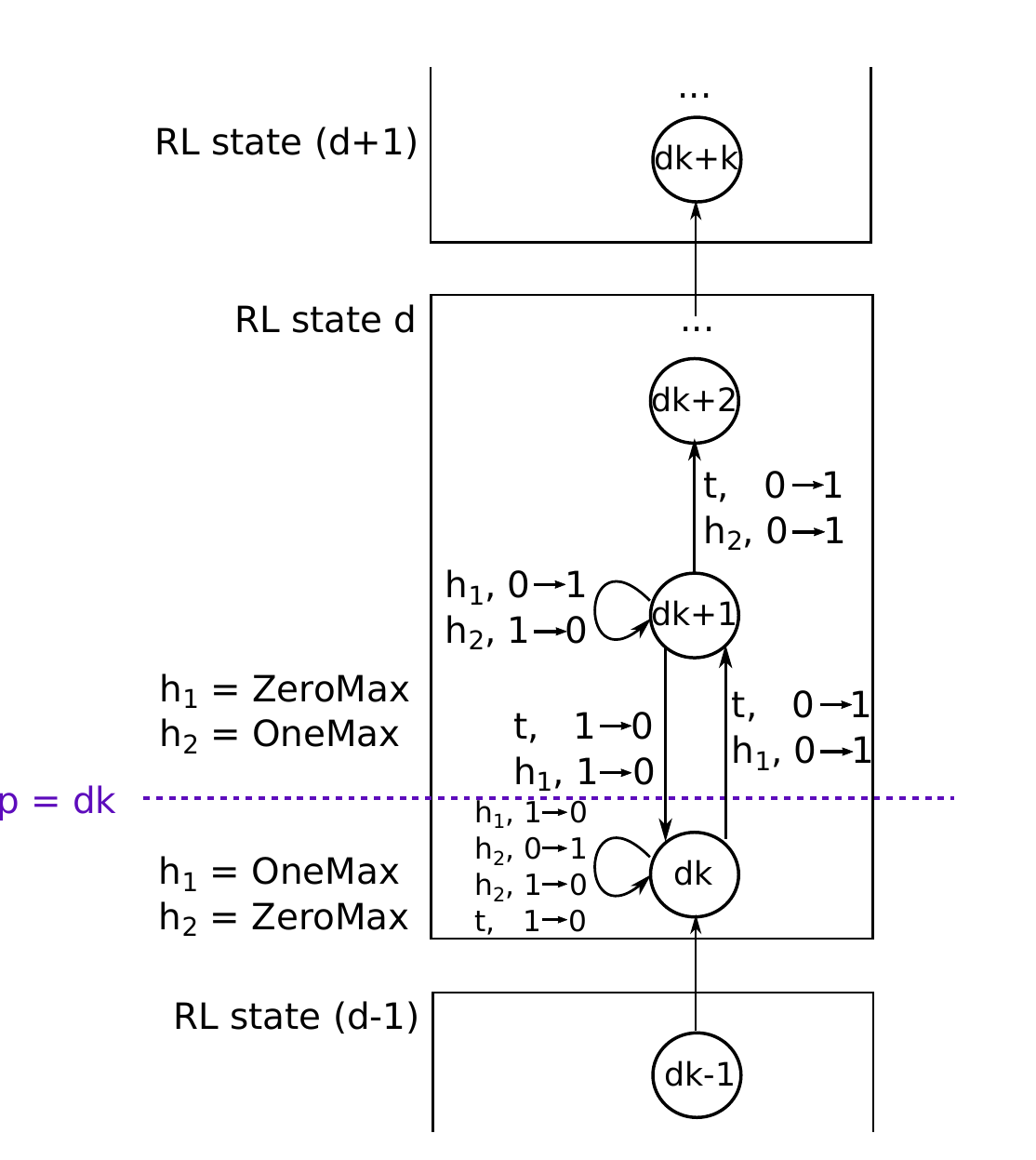}
\caption{Markov chain for the EA+RL modification without learning on mistakes on \textsc{XdivK}}
\label{markov_chain2}
\end{center}
\end{figure}

The Markov chain for the \textsc{XdivK} problem is shown in the Fig.~\ref{markov_chain2}. The labels on transitions have the format $\langle \text{F, M} 
\rangle$, where F is a fitness function that can be chosen for this transition, M is the corresponding effect 
of mutation. A transition probability is computed as the sum for all $f \in F$ of product of probabilities of selection $f$ and the corresponding mutation $m$. 
 
Further we describe each Markov state and explain why the transitions have such labels.

Consider the number of ones equal to $dk$, where $d$ is a constant. So the agent is in the RL state $d$ and the Markov state is $dk$. Since the agent has no 
experience in the state $d$, the objectives are selected equiprobably. If the agent selects the target objective or an objective which is currently equal to 
\textsc{OneMax}, and the mutation operator inverts 1-bit, the new individual has $dk-1$ 1-bits and is worse than the current individual according to the 
selected objective. 
So 
the new individual will not be accepted by EA. The same situation occurs, if the selected objective is equal to \textsc{ZeroMax} and the mutation operator 
inverts 0-bit. 
In the case of selection of the \textsc{ZeroMax} objective and inversion of 1-bit, the new individual is better than the current one according to the selected 
objective. However, 
the target objective value of the new individual is less than the target objective 
value of the current individual. Therefore, the new individual is not selected for the next generation. If \textsc{OneMax} or the target objective is selected 
and 0-bit is inverted, the new individual is accepted.  

The transitions in Markov states $dk$ and $dk+1$ differ from each other when 1-bit is inverted and the agent selects the target objective or the objective 
which is equal to 
\textsc{ZeroMax}. In this case, the new individual is equal to or better than the current one according to the selected objective. So the new individual 
is accepted. However, the new individual contains less 1-bits than the current one, so the algorithm moves to the $dk$ Markov state. Transitions in the states 
$dk+2, \ldots, dk+k-1$ are the same as transitions in the state $dk+1$. From the Markov chain of the algorithm, we can see, that transitions and, as a 
consequence, performance of the algorithm do not depend on the number and positions of switch points.  

\begin{figure}[h]
\begin{center}
\includegraphics[scale=0.8]{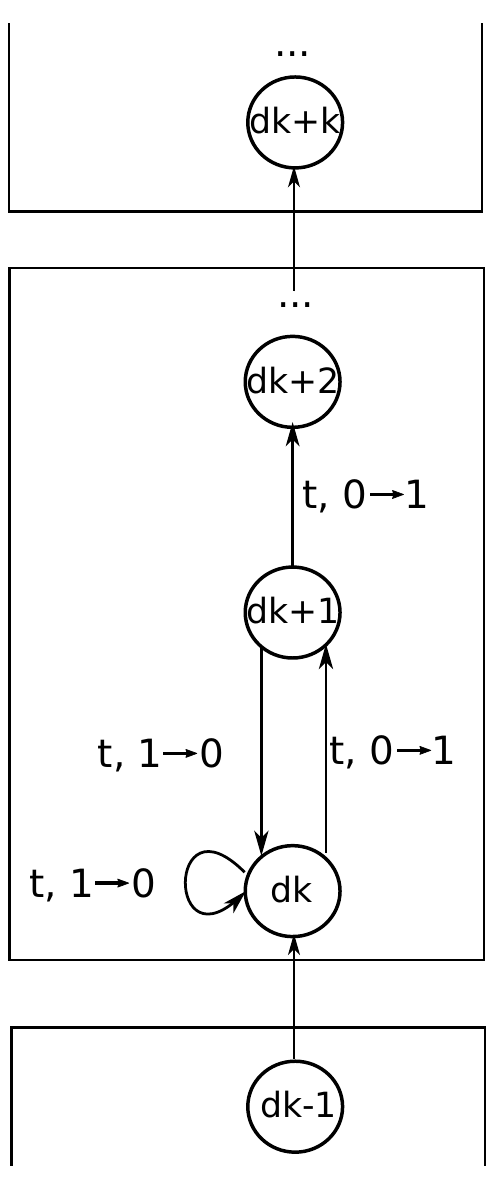}
\caption{Markov chain for RLS without auxiliary objectives on \textsc{XdivK}}
\label{markov_chain1}
\end{center}
\end{figure}

To analyse the running time of the considered algorithm, we also need to construct Markov chain for RLS without auxiliary objectives 
(see Fig.~\ref{markov_chain1}). This Markov chain is
constructed analogically to the Markov chain described above.

The expected running time of the EA+RL modification without learning on mistakes for \textsc{XdivK} with non-stationary objectives is equal to the number of 
fitness function evaluations needed to get from the Markov state $0$ to the Markov state $n$. Each transition in the Markov chain corresponds to one fitness 
function evaluation of the mutated individual. So the expected running time is equal to the number of transitions in the Markov chain. Denote the expected 
running time of the algorithm as $T(n)$: 
\begin{equation}
T(n) = \sum \limits_{i=0}^{n-1} E(i \rightarrow i+1), 
\label{teq}
\end{equation} 
where $E(i \rightarrow i+1)$ is the expected number of transitions needed to reach the Markov state $i+1$ from the state $i$. 

Consider two cases for the state $i$. The first is $i = dk$, where $d$ is a constant. The expected number of transitions needed to reach the state $dk+1$ from 
the state $dk$ is evaluated as $z_{dk} = E(dk \rightarrow dk+1)$: 
\begin{equation}
z_{dk} = \frac{2}{3}\cdot \frac{(n-dk)}{n} \cdot 1 + (\frac{2}{3}\cdot\frac{dk}{n} + \frac{1}{3}) \cdot (1+z_{dk})\label{eq1}
\end{equation}

From \eqref{eq1} we obtain that $z_{dk}$ is evaluated as:
\begin{equation}
z_{dk} = \frac{3n}{2(n-dk)}
\label{eqzdk}
\end{equation}

The second case is $i = dk+t$, where $ 1 \leq t \leq k-1 $. The expected number of transitions needed to reach the state $dk+t+1$ from the state $dk+t$ is 
evaluated 
as $z_{dk+t} = E(dk+t \rightarrow dk+t+1)$:

\begin{gather}
 z_{dk+t} = \frac{2(n-dk-t)}{3n} + \frac{2(dk+t)}{3n} \cdot (1+z_{dk+t-1} +z_{dk+t}) + 
            \nonumber\\ +
            (\frac{dk+t}{3n} + \frac{n-dk-t}{3n}) \cdot (1+z_{dk+t})
\label{eq2}
\end{gather}

From \eqref{eq2} we obtain that $z_{dk+t}$ is evaluated as:
\begin{equation}
z_{dk+t} = z_{dk+t-1}\cdot\frac{dk+t}{n-dk-t} + \frac{3n}{2(n-dk-t)}
\label{eqzdkt}
\end{equation}

To estimate the efficiency of the analysed EA+RL modification, let us calculate the expected running time of RLS without auxiliary objectives. The evaluation 
approach is similar to the one presented 
above for the EA+RL modification. The total running time is calculated by \eqref{teq}. Analogically, we consider two cases: $i=dk$ and $i=dk+t$.   
The expected number of transitions needed to reach the state $dk+1$ from the state $dk$ is evaluated as $a_{dk} = E(dk \rightarrow dk+1)$: 
\begin{equation} 
a_{dk} = \frac{(n-dk)}{n} \cdot 1 + \frac{dk}{n} \cdot (1+a_{dk}) \label{eq3}
\end{equation}

From \eqref{eq3} we obtain that $a_{dk}$ is evaluated as:
\begin{equation}
a_{dk} = \frac{n}{(n-dk)}
\label{eqadk}
\end{equation}

The expected value of transitions needed to reach the state $dk+t+1$ from the state $dk+t$ is evaluated as $a_{dk+t} = E(dk+t \rightarrow dk+t+1)$:

\begin{equation}
a_{dk+t} = \frac{(n-dk-t)}{n} \cdot 1 + \frac{dk+t}{n}\cdot (1+a_{dk+t-1} +a_{dk+t})  
\label{eq4}
\end{equation}

From \eqref{eq4} we obtain that $a_{dk+t}$ is evaluated as:
\begin{equation}
a_{dk+t} = a_{dk+t-1}\cdot\frac{dk+t}{n-dk-t} + \frac{n}{(n-dk-t)}
\label{eqadkt}
\end{equation}

From~\eqref{eqzdk} and~\eqref{eqadk} we obtain that $z_{dk} = \frac{3}{2} a_{dk}$. From the equations~\eqref{teq},~\eqref{eqzdkt},~\eqref{eqadkt} 
using mathematical induction we obtain that the running time of EA+RL modification without learning on mistakes for \textsc{XdivK} with non-stationary 
objectives is 1.5 times greater than the running time of RLS without auxiliary objective. 
Therefore, the EA+RL modification without learning on mistakes has asymptotically the same running time as RLS, which is bounded by $\Omega(n^{k})$ and 
$O(n^{k+1})$~\cite{xdivk}. This shows that the EA+RL modification without learning on mistakes can deal with non-stationary auxiliary objectives unlike EA+RL 
does.

\section{Experimental Analysis of Modified EA+RL}\label{earl-average} 
In this section we experimentally evaluate efficiency of the both proposed EA+RL modifications on the optimization problems defined in 
Section~\ref{modelproblems}. 
\subsection{Description of experiments}
We analysed the two versions of the proposed modification of EA+RL and the EA+RL method on \omdf, \textsc{XdivK} and 
\textsc{LeadingOnes}. The non-stationary objectives described in~\eqref{h} were used for all the problems. For \textsc{XdivK} we analysed two cases of the 
switch point position. 
The first case is the worst case~\cite{petrova-mendel2016}, when the switch point is in the end of optimization, $p = n-k+1$. In the second case, the switch 
point is in the middle of optimization, $p = n/2$. For each algorithm we analysed two state definitions: the single state and the target state. Also we 
studied applying of 
$\varepsilon$-greedy strategy. In this strategy the agent selects the objective with the maximum expected reward with probability $1- \varepsilon$ and with 
probability $\varepsilon$ the agent selects a random objective. This strategy gives the agent an opportunity to select the objective which was inefficient but 
became efficient after the switch point. 

The obtained numbers of fitness function evaluations needed to reach the optimum were averaged by 1000 runs. We used Q-learning algorithm 
with the same parameters as in~\cite{petrova-mendel2016}. The learning rate was set to $\alpha = 0.5$ and the discount factor was equal to $\gamma= 0.5$. The 
$\varepsilon$-greedy strategy was used with $\varepsilon = 0.1$.

\subsection{Discussion of experiment results}
Table~\ref{results} presents the results of the experiments. The first column contains parameter values for the considered problems. The next column 
contains results of RLS without auxiliary objectives. The next three columns 
correspond to results of EA+RL modification with learning on mistakes (modified EA+RL, learning). The following three columns correspond to the results of 
EA+RL 
modification without learning on mistakes (modified EA+RL, no learning). The last three columns correspond to the results of the EA+RL method (EA+RL). 
Each 
algorithm was analysed on the single state (ss, $\varepsilon=0$ and ss, $\varepsilon=0.1$) and the target state (ts, $\varepsilon=0$ and ts, 
$\varepsilon=0.1$). 
None of the algorithms reached the optimum using the single state and $\varepsilon=0$. So these results are not presented. When the optimum was not reached 
within $10^9$ iterations the corresponding result is marked as "inf".

\begin{table*}[h!]
\centering
\caption{Averaged number of runs needed to reach the global optimum}
\label{results}
\begin{tabular}{|l|l|lll|lll|lll|}
\hline
& & \multicolumn{3}{|c|}{modified EARL, learning}   & \multicolumn{3}{c|}{modified EARL, no learning} & \multicolumn{3}{c|}{EARL}   \\ \hline
Parameters & RLS & ss, $\varepsilon=0.1$ & ts, $\varepsilon=0$ & ts, $\varepsilon=0.1$ & ss, $\varepsilon=0.1$ & ts, $\varepsilon=0$ &  ts, $\varepsilon=0.1$ & 
ss, $\varepsilon=0.1$ & ts, $\varepsilon=0$ & ts, $\varepsilon=0.1$ \\ \hline
n  & \multicolumn{10}{c|}{LeadingOnes}  \\ \hline
1       & \cellcolor{res}1.46 & \cellcolor{res2}1.65 & 1.72 & 1.76 & 1.77 & 1.72 & 1.77 &  inf & inf & inf    \\
11  & \cellcolor{res}$6.20\cdot 10^1$ & $7.00\cdot 10^1$ & \cellcolor{res2}$6.63\cdot 10^1$ & $7.05\cdot 10^1$ & $6.97\cdot 10^1$ & $9.07\cdot 10^1$ & 
$9.20\cdot 10^1$ &  inf & inf & inf \\
21  & $2.21\cdot 10^2$ & \cellcolor{res2}$2.09\cdot 10^2$ & \cellcolor{res}$2.06\cdot 10^2$ & $2.15\cdot 10^2$ & $2.54\cdot 10^2$ & $3.27\cdot 10^2$ & 
$3.31\cdot 10^2$ &  inf & inf & inf \\
31  & $4.76\cdot 10^2$ & \cellcolor{res}$3.97\cdot 10^2$ & \cellcolor{res2}$4.22\cdot 10^2$ & $4.43\cdot 10^2$ & $5.67\cdot 10^2$ & $7.24\cdot 10^2$ & 
$7.23\cdot 10^2$ &  inf & inf & inf \\
41  & $8.45\cdot 10^2$ & \cellcolor{res}$6.71\cdot 10^2$ & \cellcolor{res2}$7.04\cdot 10^2$ & $7.42\cdot 10^2$ & $1.02\cdot 10^3$ & $1.26\cdot 10^3$ & 
$1.25\cdot 10^3$ &  inf & inf & inf \\
51  & $1.29\cdot 10^3$ & \cellcolor{res}$9.39\cdot 10^2$ & \cellcolor{res2}$1.06\cdot 10^3$ & $1.13\cdot 10^3$ & $1.59\cdot 10^3$ & $1.96\cdot 10^3$ & 
$1.94\cdot 10^3$ &  inf & inf & inf \\
61  & $1.86\cdot 10^3$ & \cellcolor{res}$1.27\cdot 10^3$ & \cellcolor{res2}$1.48\cdot 10^3$ & $1.58\cdot 10^3$ & $2.36\cdot 10^3$ & $2.78\cdot 10^3$ & 
$2.77\cdot 10^3$ &  inf & inf & inf \\
71  & $2.53\cdot 10^3$ & \cellcolor{res}$1.57\cdot 10^3$ & \cellcolor{res2}$1.96\cdot 10^3$ & $2.10\cdot 10^3$ & $3.23\cdot 10^3$ & $3.79\cdot 10^3$ & 
$3.79\cdot 10^3$ &  inf & inf & inf \\
81  & $3.28\cdot 10^3$ & \cellcolor{res}$1.92\cdot 10^3$ & \cellcolor{res2}$2.52\cdot 10^3$ & $2.65\cdot 10^3$ & $4.28\cdot 10^3$ & $4.94\cdot 10^3$ & 
$4.87\cdot 10^3$ &  inf & inf & inf \\
91  & $4.15\cdot 10^3$ & \cellcolor{res}$2.35\cdot 10^3$ & \cellcolor{res2}$3.16\cdot 10^3$ & $3.35\cdot 10^3$ & $5.51\cdot 10^3$ & $6.16\cdot 10^3$ & 
$6.14\cdot 10^3$ &  inf & inf & inf \\
101 & $5.07\cdot 10^3$ & \cellcolor{res}$2.78\cdot 10^3$ & \cellcolor{res2}$3.81\cdot 10^3$ & $4.04\cdot 10^3$ & $6.85\cdot 10^3$ & $7.63\cdot 10^3$ & 
$7.65\cdot 10^3$ &  inf & inf & inf \\
111 & $6.11\cdot 10^3$ & \cellcolor{res}$3.11\cdot 10^3$ & \cellcolor{res2}$4.60\cdot 10^3$ & $4.92\cdot 10^3$ & $8.38\cdot 10^3$ & $9.31\cdot 10^3$ & 
$9.19\cdot 10^3$ &  inf & inf & inf \\
121 & $7.33\cdot 10^3$ & \cellcolor{res}$3.68\cdot 10^3$ & \cellcolor{res2}$5.43\cdot 10^3$ & $5.79\cdot 10^3$ & $1.00\cdot 10^4$ & $1.10\cdot 10^4$ & 
$1.10\cdot 10^4$ &  inf & inf & inf \\
131 & $8.60\cdot 10^3$ & \cellcolor{res}$4.17\cdot 10^3$ & \cellcolor{res2}$6.36\cdot 10^3$ & $6.73\cdot 10^3$ & $1.17\cdot 10^4$ & $1.29\cdot 10^4$ & 
$1.28\cdot 10^4$ &  inf & inf & inf \\
141 & $1.00\cdot 10^4$ & \cellcolor{res}$4.61\cdot 10^3$ & \cellcolor{res2}$7.20\cdot 10^3$ & $7.80\cdot 10^3$ & $1.36\cdot 10^4$ & $1.49\cdot 10^4$ & 
$1.49\cdot 10^4$ &  inf & inf & inf \\
151 & $1.13\cdot 10^4$ & \cellcolor{res}$5.08\cdot 10^3$ & \cellcolor{res2}$8.33\cdot 10^3$ & $8.90\cdot 10^3$ & $1.57\cdot 10^4$ & $1.72\cdot 10^4$ & 
$1.72\cdot 10^4$ &  inf & inf & inf \\
161 & $1.30\cdot 10^4$ & \cellcolor{res}$5.44\cdot 10^3$ & \cellcolor{res2}$9.39\cdot 10^3$ & $1.01\cdot 10^4$ & $1.81\cdot 10^4$ & $1.94\cdot 10^4$ & 
$1.96\cdot 10^4$ &  inf & inf & inf \\
171 & $1.45\cdot 10^4$ & \cellcolor{res}$6.04\cdot 10^3$ & \cellcolor{res2}$1.06\cdot 10^4$ & $1.13\cdot 10^4$ & $2.05\cdot 10^4$ & $2.18\cdot 10^4$ & 
$2.19\cdot 10^4$ &  inf & inf & inf \\
181 & $1.65\cdot 10^4$ & \cellcolor{res}$6.60\cdot 10^3$ & \cellcolor{res2}$1.18\cdot 10^4$ & $1.27\cdot 10^4$ & $2.29\cdot 10^4$ & $2.47\cdot 10^4$ & 
$2.46\cdot 10^4$ &  inf & inf & inf \\
191 & $1.81\cdot 10^4$ & \cellcolor{res}$7.28\cdot 10^3$ & \cellcolor{res2}$1.33\cdot 10^4$ & $1.41\cdot 10^4$ & $2.58\cdot 10^4$ & $2.73\cdot 10^4$ & 
$2.73\cdot 10^4$ &  inf & inf & inf \\

\hline
n, d  &  \multicolumn{10}{c|}{OMd}  \\ \hline
100, 50& \cellcolor{res}$4.51\cdot 10^2$ & \cellcolor{res2}$4.93\cdot 10^2$ & $5.65\cdot 10^2$ & $5.69\cdot 10^2$ & $6.49\cdot 10^2$ & $6.75\cdot 10^2$ & 
$6.81\cdot 10^2$ & $\infty$ & 
$\infty$ & $\infty$\\
200, 100& \cellcolor{res}$1.04\cdot 10^3$ & \cellcolor{res2}$1.09\cdot 10^3$ & $1.26\cdot 10^3$ & $1.31\cdot 10^3$ & $1.47\cdot 10^3$ & $1.55\cdot 10^3$ & 
$1.57\cdot 10^3$ & $\infty$ & 
$\infty$ & $\infty$\\
300, 150& \cellcolor{res}$1.72\cdot 10^3$ & \cellcolor{res2}$1.74\cdot 10^3$ & $2.03\cdot 10^3$ & $2.05\cdot 10^3$ & $2.40\cdot 10^3$ & $2.51\cdot 10^3$ & 
$2.51\cdot 10^3$ & $\infty$ & 
$\infty$ & $\infty$\\
400, 200& \cellcolor{res}$2.43\cdot 10^3$ & \cellcolor{res2}$2.43\cdot 10^3$ & $2.80\cdot 10^3$ & $2.90\cdot 10^3$ & $3.42\cdot 10^3$ & $3.56\cdot 10^3$ & 
$3.53\cdot 10^3$ & $\infty$ & 
$\infty$ & $\infty$\\
500, 250& \cellcolor{res}$3.12\cdot 10^3$ & \cellcolor{res2}$3.16\cdot 10^3$ & $3.65\cdot 10^3$ & $3.72\cdot 10^3$ & $4.34\cdot 10^3$ & $4.58\cdot 10^3$ & 
$4.60\cdot 10^3$ & $\infty$ & 
$\infty$ & $\infty$\\

\hline
n, k & \multicolumn{10}{c|}{XdivK, switch point in the end}                 \\ \hline
40, 2  & \cellcolor{res}$1.19\cdot 10^3$ & $3.20\cdot 10^3$ & $3.67\cdot 10^3$ & $3.25\cdot 10^3$ & $2.52\cdot 10^3$ & \cellcolor{res2}$1.73\cdot 10^3$ & 
$1.77\cdot 10^3$ &  inf &$3.72\cdot 
10^3$ & $2.28\cdot 10^4$\\
48, 2  & \cellcolor{res}$1.70\cdot 10^3$ & $4.46\cdot 10^3$ & $5.05\cdot 10^3$ & $4.31\cdot 10^3$ & $3.73\cdot 10^3$ & \cellcolor{res2}$2.37\cdot 10^3$ & 
$2.49\cdot 10^3$ &  inf &$5.23\cdot 
10^3$ & $4.86\cdot 10^4$\\
56, 2  & \cellcolor{res}$2.30\cdot 10^3$ & $5.95\cdot 10^3$ & $6.69\cdot 10^3$ & $6.45\cdot 10^3$ & $4.92\cdot 10^3$ & $3.50\cdot 10^3$ & 
\cellcolor{res2}$3.46\cdot 10^3$ &  inf &$6.81\cdot 
10^3$ & $1.59\cdot 10^5$\\
64, 2  & \cellcolor{res}$2.98\cdot 10^3$ & $7.86\cdot 10^3$ & $8.48\cdot 10^3$ & $7.78\cdot 10^3$ & $6.31\cdot 10^3$ & \cellcolor{res2}$4.47\cdot 10^3$ & 
$4.55\cdot 10^3$ &  inf &$8.91\cdot 
10^3$ & $3.75\cdot 10^5$\\
72, 2  & \cellcolor{res}$3.76\cdot 10^3$ & $9.42\cdot 10^3$ & $1.12\cdot 10^4$ & $1.03\cdot 10^4$ & $8.05\cdot 10^3$ & $5.60\cdot 10^3$ & 
\cellcolor{res2}$5.47\cdot 10^3$ &  inf &$1.11\cdot 
10^4$ & $1.38\cdot 10^6$\\
80, 2  & \cellcolor{res}$4.62\cdot 10^3$ & $1.25\cdot 10^4$ & $1.33\cdot 10^4$ & $1.30\cdot 10^4$ & $9.38\cdot 10^3$ & \cellcolor{res2}$6.74\cdot 10^3$ & 
$7.07\cdot 10^3$ &  inf &$1.41\cdot 
10^4$ & $3.56\cdot 10^6$\\
60, 3  & \cellcolor{res}$3.94\cdot 10^4$ & $2.44\cdot 10^5$ & $2.79\cdot 10^5$ & $2.53\cdot 10^5$ & $7.10\cdot 10^4$ & \cellcolor{res2}$5.82\cdot 10^4$ & 
$5.95\cdot 10^4$ &  inf &$2.95\cdot 
10^5$ &  $3.16\cdot 10^7$ \\
72, 3  & \cellcolor{res}$6.79\cdot 10^4$ & $4.18\cdot 10^5$ & $4.93\cdot 10^5$ & $4.19\cdot 10^5$ & $1.15\cdot 10^5$ & \cellcolor{res2}$1.03\cdot 10^5$ & 
$1.03\cdot 10^5$ &  inf &$4.98\cdot 
10^5$ &  $3.30 \cdot 10^8$\\
84, 3  & \cellcolor{res}$1.08\cdot 10^5$ & $6.57\cdot 10^5$ & $7.69\cdot 10^5$ & $6.55\cdot 10^5$ & $1.72\cdot 10^5$ & \cellcolor{res2}$1.61\cdot 10^5$ & 
$1.64\cdot 10^5$ &  inf &$7.81\cdot 
10^5$ &  inf     \\
96, 3  & \cellcolor{res}$1.60\cdot 10^5$ & $9.91\cdot 10^5$ & $1.21\cdot 10^6$ & $9.97\cdot 10^5$ & $2.59\cdot 10^5$ & \cellcolor{res2}$2.39\cdot 10^5$ & 
$2.44\cdot 10^5$ &  inf &$1.05\cdot 
10^6$ &  inf     \\
108, 3 & \cellcolor{res}$2.28\cdot 10^5$ & $1.34\cdot 10^6$ & $1.75\cdot 10^6$ & $1.45\cdot 10^6$ & $3.49\cdot 10^5$ & \cellcolor{res2}$3.34\cdot 10^5$ & 
$3.34\cdot 10^5$ &  inf &$1.63\cdot 
10^6$ &  inf     \\
120, 3 & \cellcolor{res}$3.12\cdot 10^5$ & $1.93\cdot 10^6$ & $2.32\cdot 10^6$ & $1.97\cdot 10^6$ & $4.87\cdot 10^5$ & \cellcolor{res2}$4.70\cdot 10^5$ & 
$4.76\cdot 10^5$ &  inf &$2.37\cdot 
10^6$ &  inf     \\
\hline
n, k  & \multicolumn{10}{c|}{XdivK, switch point in the middle}    \\ \hline
40, 2  & \cellcolor{res2}$1.19\cdot 10^3$ & $9.06\cdot 10^2$ & $6.63\cdot 10^2$ & $7.13\cdot 10^2$ & $2.46\cdot 10^3$ & $1.73\cdot 10^3$ & $1.76\cdot 10^3$ &  
inf     
&\cellcolor{res}$8.28\cdot 10^2$ & $3.57\cdot 10^3$\\
48, 2  & $1.70\cdot 10^3$ & $1.22\cdot 10^3$ & $9.76\cdot 10^2$ & \cellcolor{res}$1.01\cdot 10^3$ & $3.41\cdot 10^3$ & $2.37\cdot 10^3$ & $2.61\cdot 10^3$ &  
inf     
&\cellcolor{res2}$1.18\cdot 10^3$ & $9.59\cdot 10^3$\\
56, 2  & $2.30\cdot 10^3$ & $1.50\cdot 10^3$ & \cellcolor{res}$1.28\cdot 10^3$ & \cellcolor{res2}$1.35\cdot 10^3$ & $4.76\cdot 10^3$ & $3.50\cdot 10^3$ & 
$3.45\cdot 10^3$ &  inf     
&$1.56\cdot 10^3$ & $2.43\cdot 10^4$\\
64, 2  & $2.98\cdot 10^3$ & $1.84\cdot 10^3$ & \cellcolor{res}$1.66\cdot 10^3$ & \cellcolor{res2}$1.77\cdot 10^3$ & $6.09\cdot 10^3$ & $4.47\cdot 10^3$ & 
$4.23\cdot 10^3$ &  inf     
&$2.04\cdot 10^3$ & $7.08\cdot 10^4$\\
72, 2  & $3.76\cdot 10^3$ & \cellcolor{res2}$2.13\cdot 10^3$ & \cellcolor{res}$2.03\cdot 10^3$ & $2.23\cdot 10^3$ & $7.61\cdot 10^3$ & $5.60\cdot 10^3$ & 
$5.41\cdot 10^3$ &  inf     
&$2.55\cdot 10^3$ & $1.91\cdot 10^5$\\
80, 2  & $4.62\cdot 10^3$ & \cellcolor{res}$2.48\cdot 10^3$ & \cellcolor{res2}$2.61\cdot 10^3$ & $2.75\cdot 10^3$ & $9.66\cdot 10^3$ & $6.74\cdot 10^3$ & 
$6.56\cdot 10^3$ &  inf     
&$3.02\cdot 10^3$ & $4.76\cdot 10^5$\\
60, 3  & $3.94\cdot 10^4$ & \cellcolor{res}$6.61\cdot 10^3$ & \cellcolor{res2}$1.06\cdot 10^4$ & $1.20\cdot 10^4$ & $7.02\cdot 10^4$ & $5.82\cdot 10^4$ & 
$5.76\cdot 10^4$ &  inf     
&$1.17\cdot 10^4$ & $1.52\cdot 10^6$\\
72, 3  & $6.79\cdot 10^4$ & \cellcolor{res}$1.12\cdot 10^4$ & \cellcolor{res2}$1.78\cdot 10^4$ & $2.11\cdot 10^4$ & $1.19\cdot 10^5$ & $1.03\cdot 10^5$ & 
$1.01\cdot 10^5$ &  inf     
&$2.00\cdot 10^4$ & $1.43\cdot 10^7$\\
84, 3  & $1.08\cdot 10^5$ & \cellcolor{res}$1.79\cdot 10^4$ & \cellcolor{res2}$3.02\cdot 10^4$ & $3.35\cdot 10^4$ & $1.73\cdot 10^5$ & $1.61\cdot 10^5$ & 
$1.64\cdot 10^5$ &  inf     
&$3.08\cdot 10^4$ & $1.57\cdot 10^8$\\
96, 3  & $1.60\cdot 10^5$ & \cellcolor{res}$3.01\cdot 10^4$ & \cellcolor{res2}$4.13\cdot 10^4$ & $5.15\cdot 10^4$ & $2.60\cdot 10^5$ & $2.39\cdot 10^5$ & 
$2.44\cdot 10^5$ &  inf     
&$4.75\cdot 10^4$ & inf   \\
108, 3 & $2.28\cdot 10^5$ & \cellcolor{res}$4.54\cdot 10^4$ & \cellcolor{res2}$5.86\cdot 10^4$ & $6.71\cdot 10^4$ & $3.74\cdot 10^5$ & $3.34\cdot 10^5$ & 
$3.43\cdot 10^5$ &  inf     
&$6.88\cdot 10^4$ & inf   \\
120, 3 & $3.12\cdot 10^5$ & \cellcolor{res}$6.46\cdot 10^4$ & \cellcolor{res2}$8.25\cdot 10^4$ & $9.37\cdot 10^4$ & $4.98\cdot 10^5$ & $4.70\cdot 10^5$ & 
$4.59\cdot 10^5$ &  inf     
&$9.32\cdot 10^4$ & inf  \\

\hline
\end{tabular}
\end{table*}

We can see from Table~\ref{results} that the modification of EA+RL with learning on mistakes using the single state and $\varepsilon = 0.1$ is the most 
efficient algorithm on \textsc{LeadingOnes}, \omdf and \textsc{XdivK} with switch point in the middle. On LeadingOnes and XdivK with switch point 
in the middle this algorithm ignores an inefficient objective and selects efficient one, so the achieved results are better than the results of RLS without 
objectives. On \omdf this modification ignores obstructive objectives and achieves the same results as RLS. On XdivK with switch point in the end, the 
best results are achieved using the modification of EA+RL without learning on mistakes. For each problem, we picked the best configuration of each algorithm and 
compared them by Mann-Whitney test 
with Bonferroni correction. The algorithms were statistically distinguishable with p-value less than 0.05. Below we analyse how different components of the 
considered 
methods influence optimization performance. More precisely, we consider influence of the best individual preservation, learning which objective is 
obstructive, state definition and $\varepsilon$-greedy strategy.

\subsubsection{Influence of learning which objective is obstructive} \label{learninginfl}
We can see from the results that the modification of EA+RL without learning is 
outperformed by the version with learning on \textsc{LeadingOnes} and \textsc{XdivK} with switch point in the middle. Therefore, 
learning on mistakes is useful because it allows the agent to remember that the objective is obstructive and not to select it further. However, on XdivK 
problem with switch point in the end, the best results are achieved using modification of EA+RL without learning on mistakes. Below we explain why learning on 
mistakes is not always efficient. 

If some objective becomes helpful, the agent will not select this objective in the same state because it obtained a negative reward for it. So there 
are two ways for the agent to learn that the obstructive objective became helpful. The first way is to select this objective with $\varepsilon$-probability and 
achieve a positive reward. The second way is to move to the new state, where the agent has not learned which objective is efficient. It is impossible when 
using the single state. Consider what actions agent should do to move to the new state if the target state is used. The only way to move 
to the new state is increasing of the target objective value. The agent could select the target objective or the objective which was efficient but became 
obstructive. So target objective value can be increased only if the new individual has a higher target objective value and the selected objective is the 
target one. 

In some optimization problems it is not always possible to increase the target objective in one iteration of the algorithm. For example, consider 
\textsc{XdivK} problem. Let the number of 1-bits be $dk$, so the RL state is $d$. To move to the state $d+1$, the algorithm needs to mutate $k$ 0-bits. Let 
switch point $p$ be equal to $dk+l$, where $0<l<k$. Then if the number of 1-bits is greater than $dk+l$, the algorithm can increase the number of 1-bits 
only if 0-bit 
is mutated and the target objective is selected. However, whatever bit is mutated and whatever objective is selected, the target objective value stays 
unchanged until an individual with $dk+k$ 1-bits is obtained. So the agent does not recognize if its action is good or bad because the reward is equal to zero. 
Therefore, to increase the target objective value 
algorithm needs a lot of steps. In addition, if the switch point is in the end of optimization, probability to mutate a 0-bit and select the target objective 
at the same time is low. The worst case is when the switch point is equal to $n-k+1$, because the agent needs to select the target objective and 
increase the number of 1-bits during $k-1$ iterations. The modification of EA+RL without learning on 
mistakes does not have this drawback, because as it was shown in theoretical analysis, it does not depend on the number of switch points and their 
positions. 

To conclude, the modification of EA+RL with learning performs better than the proposed modification without learning, except 
the case when the switch point occurs at the time when it is hard to generate good solutions (stagnation).

\subsubsection{Influence of best individual preservation}
Thanks to preserving of individual with the best target value, the proposed modification of EA+RL achieves the optimum on \textsc{LeadingOnes} and \omdf 
problems unlike the EA+RL method does. However, we can see that on XdivK with switch point in the middle EA+RL achieves better results despite the possibility 
to lose the best individual. It is explained by the fact that EA+RL learns which objective is obstructive and for this problem it is more important than 
preserving of the best individual. Inverse situation is observed for XdivK with switch point in the end where learning does not help to achieve the optimum 
faster even in case of preserving of the best individual (see section~\ref{learninginfl}). 

To conclude, preservation of the best individual improves the EA+RL method. However, learning which objective is efficient has a greater impact on the 
algorithm performance.   

\subsubsection{Influence of state definition} 
To begin with, let us separately consider RL efficiency when using the target state and the single state. Then we will discuss how state definition 
influence the performance 
of two proposed modifications of EA+RL.

Consider the target state. The algorithm moves to the new state if the target objective value is increased and, as follows, a positive 
reward is obtained. If the target objective value can not be decreased, the algorithm never returns to the state where it achieved a positive reward.  
Therefore, the agent never knows which action is efficient if the target state is used. 

The single state allows to learn which objective is helpful contrary to the target state. However, when the objective which was efficient becomes obstructive, 
the agent continues selection of this objective. Re-learning that the objective which was helpful became obstructive when using the single state is more 
difficult than if the target state is used (see Section~\ref{learninginfl}). None of the algorithms using the single 
state with 
$\varepsilon = 0$ reached the optimum. So without $\varepsilon$-greedy strategy the agent could not re-learn. Also EA+RL without preservation of the best 
obtained individual does not achieve the optimum using the single state. 

Consider influence of state definition in the EA+RL modification with learning. When solving \textsc{LeadingOnes}, the 
positive effect of ability to learn which objective is efficient is more important than the negative effect of difficult 
re-learning. We can see different influences of 
these negative and positive effects when solving the XdivK problem with switch point in the middle. When $k$ becomes bigger, it is more important to select an 
efficient 
objective during $k$ iterations where agent does not achieve reward and could not define which objective is better. So we can see that for $k=3$ the single 
state is much better than the target state. On XdivK problem with switch point in the end, the results obtained using the single state and the target state 
with 
$\varepsilon = 0.1$ are the same. These results are better than the results obtained using the target state with $\varepsilon = 0$. Therefore, the impact of 
$\varepsilon$ is greater than the impact of state definition. 

In the EA+RL modification without learning the agent does not achieve negative reward so re-learning 
when using the single state is harder. Re-learning can occur only if a positive reward is obtained when a better solution is generated. In XdivK, 
many iterations are needed to increase the target objective value (see Section~\ref{learninginfl}). Therefore, the results on XdivK for the target state are 
better than the results 
for the single state. In \textsc{LeadingOnes}, generating of an individual with a higher target objective value is not so difficult. Therefore, using the single 
state the algorithm obtains better results than using the target state.

To conclude, in the modification of EA+RL with learning on mistakes the single state is better in the most cases. In the modification of EA+RL without 
learning the single state is worse than the target only on XdivK where re-learning is difficult. Also we can note that EA+RL does not achieve the 
optimum using the single state.

\subsubsection{Influence of $\varepsilon$-greedy strategy}
Consider influence of $\varepsilon$ value on performance of the considered algorithms when using the target state. In this state the agent can only learn 
that an objective is obstructive, if a negative reward is achieved after applying this objective. So this objective will not be selected in the same state even 
if it will become helpful. Non-zero $\varepsilon$ allows to select this objective. In the modification of EA+RL without 
learning on mistakes the agent can obtain only non-negative reward. So the situation described above is impossible in this modification. Therefore, 
$\varepsilon$ value does not influence the efficiency of this modification. In the 
modification of EA+RL with learning on mistakes non-zero $\varepsilon$ is helpful only on XdivK problem with switch point in the 
end. On the other problems the results obtained with non-zero $\varepsilon$ are worse than the results obtained with $\varepsilon = 0$. 

When using the single state, $\varepsilon$ allows to select the objective which was obstructive, as if the 
target state was used. However, if the single state is used, the agent learns not only that an objective is inefficient, but also that an objective is 
efficient. 
Therefore, $\varepsilon$ also allows not to select the objective which was efficient but became obstructive. This results in reaching the optimum when using 
non-zero $\varepsilon$ in the single state.       

To conclude, if the single state is used, $\varepsilon$ value have to be non-zero. If the target state is used, non-zero $\varepsilon$ 
allows to achieve better results only if re-learning is very difficult, such as in XdivK with switch point in the end.

\section{Conclusion}\label{concl}
We proposed a modification of the EA+RL method which preserves the best found solution. We considered two versions of the proposed modification. 
In the first version, called the modification of EA+RL without learning on mistakes, the RL agent learns only when the algorithm finds a better solution. In 
the 
second version, called the modification of EA+RL with learning on mistakes, the RL agent also learns when the algorithm obtains an inefficient solution. 

We considered two auxiliary objectives which change their efficiency at switch point.
We experimentally analysed the two proposed modifications and the EA+RL method on \omdf, \textsc{LeadingOnes}, \textsc{XdivK} with switch point in the middle 
of optimization and \textsc{XdivK} with switch point in the end. Two RL state definitions were considered: the single state and the target state. Also we 
considered how $\varepsilon$-greedy exploration strategy influence the performance of the algorithm.

The both proposed modifications reached the optimum on \omdf and \textsc{LeadingOnes} unlike the EA+RL method did.
The modification of EA+RL with learning using the single state and $\varepsilon = 0.1$ achieved the best results among the considered objective selection 
methods 
on all 
problems, except the \textsc{XdivK} problem with switch point in the end. On \textsc{LeadingOnes} and \textsc{XdivK} with switch point in the middle this 
algorithm was able to select an  efficient objective and obtain better results than RLS. On \omdf without helpful objectives, this modification 
ignored obstructive objectives and achieved the same results as RLS. Therefore, keeping the best individual and using $\varepsilon$-greedy exploration in the 
single 
state seems to be the most promising reinforcement based objective selection approach. 


We theoretically proved that the lower and upper bounds on the running time of the modification of EA+RL without learning on mistakes on the \textsc{XdivK} 
problem 
are $\Omega(n^{k})$ and $O(n^{k+1})$ correspondingly. The asymptotic of RLS on \textsc{XdivK} without auxiliary objectives is the same. This means that the 
modification of EA+RL without learning on mistakes on the \textsc{XdivK} problem ignores the objective which is currently obstructive.
Also we proved that performance of this modification is independent of the number of switch points and their positions, while performance of 
the modification with learning depends on these factors. Particularly, the modification without learning achieves the best results on 
XdivK with switch point in the end. This is an especially difficult case because increasing of the target objective value in the end of optimization needs a 
lot of iterations.

\section{Acknowledgments}
This work was supported by RFBR according to the research project No. 16-31-00380 mol\_a.

\bibliographystyle{IEEEtran}
\bibliography{../../../bibliography}

\end{document}